\pdfoutput=1
\documentclass[pmlr,twocolumn,10pt]{jmlr}

\usepackage{booktabs}
\usepackage{amssymb}
\usepackage{xurl}
\providecommand{\propositionref}[1]{Proposition~\ref{#1}}

\jmlrproceedings{Preprint}{Preprint}
\jmlrvolume{}
\jmlrpages{}
\jmlryear{}
\jmlrworkshop{}
\makeatletter\renewcommand*{\@titlefoot}{}\makeatother

\title[Window-Aware Scheduling under Platform Delivery Constraints]{Window-Aware Scheduling for Health Messaging under Platform Delivery Constraints: Evidence from an AI-Powered WhatsApp Maternal and Infant Health Chatbot in Nigeria}

\author[Adepoju Jeremiah Moyondafoluwa et al.]{%
\Name{Adepoju Jeremiah Moyondafoluwa}\quad
\Name{Omolaiye Samuel}\quad
\Name{Adewuyi Thanni}\\
\Name{Okenla AbdulQahhar Abimbola}\quad
\Name{Lily-Aimas Love}\quad
\Name{Adereni Abiodun}\\[2pt]
\addr HelpMum Africa}

\hypersetup{pdftitle={Window-Aware Scheduling for Health Messaging under Platform Delivery Constraints: Evidence from an AI-Powered WhatsApp Maternal and Infant Health Chatbot in Nigeria},
  pdfauthor={Adepoju Jeremiah Moyondafoluwa, Omolaiye Samuel, Adewuyi Thanni, Okenla AbdulQahhar Abimbola, Lily-Aimas Love, Adereni Abiodun},
  pdfsubject={Preprint}}

\begin{document}
\maketitle

\begin{abstract}
WhatsApp lets a business send free-form messages only within 24 hours of a user's last message; outside this \emph{customer service window}, only pre-approved templates are delivered. Health chatbots inherit this rule, but its cost is rarely measured. We analyse 98,607 messages sent to 3,696 mothers by a WhatsApp maternal health chatbot in Nigeria, with a largely separate export as replication. No free-form message failed with the window open (0 of 35,500), and failure jumped by 93.0 percentage points (95\% CI 87.0 to 99.1) at exactly 24 hours. Two flows produced 66.1\% of failures, including an offer to speak with a medical professional that, as built, reached 0.3 in 100 mothers whose window had closed. We formalise window-aware scheduling as a decision problem with an action of known zero value. Model-based offline evaluation, with bounds for unconfirmed delivery, estimates that sending the offer as an approved template would reach 81.5 to 86.8 in 100 such mothers, against 16.6 for waiting and 11.6 for a reply prompt. In a calibrated simulation, tabular Q-learning converges to the same template rule and does not beat it in any of 13 stress tests, and opting out does not rise with message volume. The main gain comes from respecting the platform rule, and it needs no learning. All outcomes are deliveries, not health outcomes.
\end{abstract}

\section{Introduction}
\label{sec:intro}

Nigeria accounted for an estimated 28.5\% of the world's maternal deaths in 2020 \citep{who2023maternal}. Mobile messaging has been tested widely as a way to reach pregnant women with health information: systematic reviews of maternal and newborn mHealth in low- and middle-income countries report improvements in some care outcomes, with evidence of mixed quality \citep{lee2016mhealth,sondaal2016assessing}, a cluster-randomised trial in Zanzibar found that text messages increased skilled attendance at delivery among urban women \citep{lund2012mobile}, and text-message trials in Kenya improved antiretroviral adherence \citep{lester2010weltel,popeleches2011mobile}. WHO guidance on digital health includes targeted communication to clients through mobile devices \citep{who2019digital}. Some now run on WhatsApp, which is widely used in health research \citep{manji2021whatsapp}, and many take the form of chatbots, a class of tools reviewed by \citet{laranjo2018conversational}.

WhatsApp imposes a delivery rule that most programmes inherit without measuring. A business may send free-form messages only within 24 hours of the user's last message. Outside this window, only templates approved in advance by Meta are delivered, and a free-form message is rejected \citep{meta_pricing}. The rule matters because many important health messages are proactive. A reminder, a follow-up or an offer to speak with a clinician is sent by a scheduler, often when the mother has not written for days. If the scheduler ignores the window, these messages fail silently: they are logged as sent and never arrive.

We measure that loss and evaluate how to schedule around it, using the logs of a WhatsApp maternal health chatbot run by a non-profit organisation in Nigeria. We treat the window as a hard, observable constraint on which actions are available. This connects a practical delivery problem to bandits with changing action sets \citep{kanade2009sleeping,kleinberg2010regret} and to adaptive interventions that depend on when a person can be reached \citep{nahumshani2018jitai,kunzler2019receptivity}. It also changes the learning problem: the platform fixes the value of one action at zero whenever the window is closed, and the window is visible before each decision.

\paragraph{Research questions.} \textbf{RQ1:} How much proactive messaging fails because of the window, and is the effect sharp enough to identify from logs alone? \textbf{RQ2:} Which flows produce the loss, and why? \textbf{RQ3:} Which window-aware policies recover it: approved templates, waiting for the mother to write, or prompting a reply that reopens the window? \textbf{RQ4:} When today's choice changes tomorrow's window, does reinforcement learning improve on simple rules?

\paragraph{Contributions.} We make four contributions. First, we measure the message loss attributable to the 24-hour window, identified by a sharp discontinuity and replicated in a second export (\sectionref{sec:res-rq1}). Second, we diagnose its source: two flows produce two thirds of failures, and one of them blocks a clinician referral by design (\sectionref{sec:res-rq2}). Third, we formulate window-aware scheduling with a known-zero action and state when waiting or a reply prompt can beat a template (\sectionref{sec:formulation}). Fourth, we evaluate policies offline by content type, with explicit delivery bounds and stated identification assumptions, and show in a calibrated simulation that Q-learning converges to the template rule (\sectionref{sec:res-rq3} and \sectionref{sec:res-rq4}).

\figureref{fig:cliff} summarises the first finding. The broader lesson is that a learning method is only as useful as the structure left after known constraints are encoded; here, encoding one platform rule removes almost all of the loss.

\begin{figure*}[t]
\floatconts{fig:cliff}{\caption{Bot-message failure by hours since the mother's last message $h$, main export (observed). (a) Failure rate by bin with Wilson 95\% intervals, among the 65,193 messages with an earlier message observed (the 33,414 without one have no $h$; \tableref{tab:cells}). (b) 2-hour bins and the local linear fit on each side of 24 hours (bandwidth 24 hours, 95\% band); the template jump is 1.0 point (CI $-0.7$ to 2.7).}}{\includegraphics[width=\textwidth]{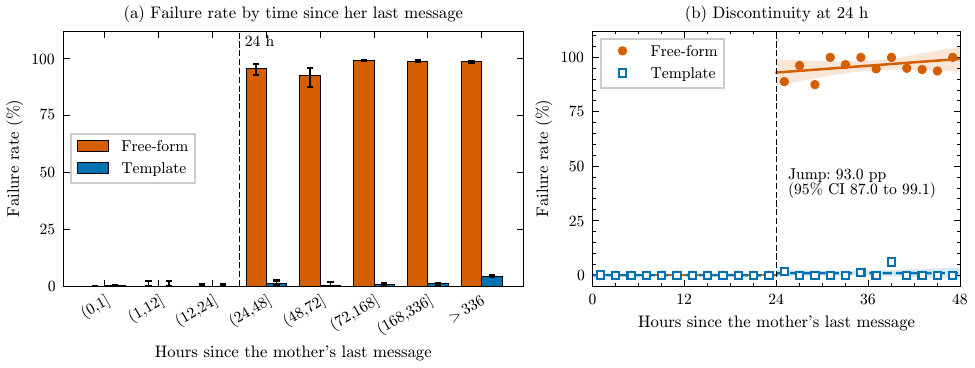}}
\end{figure*}

\section{Related Work}
\label{sec:related}

\paragraph{Maternal messaging and reach.} Large programmes show that reach often limits impact. India's Kilkari voice service lost much of its intended exposure to unanswered and short calls \citep{bashingwa2021kilkari}, and South Africa's MomConnect documented gaps in receipt and engagement \citep{barron2018momconnect,lefevre2018unpacking}. In Nigeria, a voice-messaging trial in Lagos found effects on antenatal care use \citep{osanyin2022effects}, and SMS tools are being piloted for breastfeeding support \citep{musti2025mwana}; in Kenya, a cluster-randomised trial tested a digital platform across the maternal and newborn care continuum \citep{vatsa2024empowering}. Engagement with digital health tools commonly declines over time \citep{eysenbach2005law}. These studies measure exposure on SMS and voice channels, not the delivery rules of a messaging platform.

\paragraph{Chatbots and WhatsApp in health services.} Reviews of conversational agents in health care describe many systems with limited or mixed evaluation \citep{laranjo2018conversational,milneives2020effectiveness}, and users can be hesitant about chatbot-led health services \citep{nadarzynski2019acceptability}. WhatsApp is widely used in health research \citep{manji2021whatsapp}. A diabetes chatbot in Cape Town reported messages ``queued up in the system and eventually dropped'' and no provider error logs \citep{mash2022great4diabetes}. A MomConnect trial protocol notes the template requirement as a design constraint \citep{idinsight2023momconnect}, and practitioners report high blocking of marketing templates and rising costs \citep{hughey2026observability,amazonbrown2026meta}. We found no peer-reviewed study that quantifies how much health messaging the 24-hour window blocks.

\paragraph{Timing, receptivity and adaptive interventions.} Work on just-in-time adaptive interventions studies when people are available and receptive \citep{nahumshani2018jitai,kunzler2019receptivity,mishra2021detecting}, and studies of phone notifications show that attention to a message depends on its content and the user's context \citep{pielot2014insitu,mehrotra2016myphone}. Micro-randomised trials estimate the effect of prompts sent at different moments \citep{klasnja2015microrandomized,klasnja2019efficacy}, and contextual bandits, a method also used for news recommendation \citep{li2010contextual}, and reinforcement learning personalise which prompt to send and when \citep{tewari2017ads,yomtov2017encouraging,liao2020personalized,trella2022designing,mandyam2024adaptive}. In maternal health, restless bandits allocate scarce health-worker calls in a large programme in India \citep{mate2022field,wang2023scalable,verma2026saheli}, and bandits have been used to choose call times \citep{lalan2024improving,dasgupta2025learning}. These studies treat availability as a probabilistic state learned from behaviour. On WhatsApp, availability for free-form messages is a hard state set by the platform and visible in the log.

\paragraph{Changing action sets and offline evaluation.} Sleeping bandits model actions available only in some rounds, revealed before each choice \citep{kanade2009sleeping,kleinberg2010regret}; bandits with knapsacks add budget constraints \citep{badanidiyuru2018bandits}; \citet{lattimore2020bandit} give an overview. Our setting is a special case in which availability is known from the window and the reward of an unavailable action is exactly zero. Off-policy evaluation by importance sampling \citep{precup2000eligibility,li2011unbiased} or doubly robust estimators \citep{dudik2011doubly,jiang2016doubly,thomas2016dataefficient,mandyam2026candor} needs logged action probabilities, and guidelines for reinforcement learning in health, motivated by applications such as sepsis treatment \citep{komorowski2018clinician}, warn that evaluation from observational logs can mislead without them \citep{gottesman2019guidelines}. Our logs have none, so, as in other off-policy work on clinical logs \citep{ji2025exosito}, we state the assumptions our model-based values need.

\paragraph{Gap and contribution.} Prior work on maternal messaging measures reach on SMS and voice, and work on timing treats availability as something to learn. We measure a reach loss caused by an observable platform rule, show that it is identified by a discontinuity in routine logs, and show that once the rule is encoded, the remaining decision is simple enough that learning does not improve on a rule.

\section{Setting, Data and Cohort}
\label{sec:data}

\subsection{The service}
The service runs in English on the WhatsApp Business Platform through a messaging provider. Mothers receive scheduled content and can ask questions at any time. The main flows are: weekly pregnancy guides, sent as approved templates; an offer to speak with a medical professional, a free-form interactive message with Yes and No buttons chained about a second after the weekly guide; a free-form error follow-up (``Please let me know if you ever have any questions''), sent about five minutes after the offer in the same weekly chain; vaccination, nutrition and broadcast templates; and free-form conversational replies. Mothers can opt out of individual streams.

\subsection{Platform rules during the study}
Throughout the study period, free-form messages could be delivered only inside the window, which opens or resets only when the user messages the business \citep{meta_pricing}. Templates could be sent at any time, subject to approval and, for marketing templates, per-user limits \citep{meta_marketing_limits}. Service conversations became free from November 2024, templates moved to per-message pricing in July 2025, and from 1 October 2026 in-window service messages became chargeable after 1,000 free per month \citep{meta_pricing}. Almost all our data predate any charge for in-window messages. Published utility rates for Nigeria differ between sources (US\$0.0101 and US\$0.0067 per message; \citealp{techpoint2026whatsapp,ominiflow2026nigeria}); we use the higher rate.

\subsection{Data and cohort}
The operator supplied two exports of conversation logs (\tableref{tab:data}). Each row holds a mother's message, a bot message or both, with timestamps, flow, message type and final status (read, delivered, sent, failed or unknown). The exports contain no demographic fields. Recorded pregnancy weeks for guide recipients have a median of 33, so most mothers in the data are late in pregnancy or have recently given birth.

\begin{table}[t]
\floatconts{tab:data}{\caption{The two exports and the analysis cohorts (observed counts). $^a$Bot-initiated free-form messages with 7 days of follow-up. $^b$Delivered (not failed) bot-initiated messages. Obs.: earlier message observed; none obs.: no earlier message observed.}}{%
\small\setlength{\tabcolsep}{4pt}
\begin{tabular}{@{}lrr@{}}
\toprule
 & Main & Replication \\
\midrule
Bot messages & 98,607 & 23,138 \\
\quad free-form & 50,574 & 12,266 \\
\quad templates & 48,033 & 10,872 \\
Messages from mothers & 29,297 & 7,586 \\
Mothers & 3,696 & 927 \\
Period & 2023 to 2026 & 2023 to 2026 \\
Mothers in both exports & 131 & 131 \\
Failed bot messages & 21.8\% & 21.0\% \\
No delivery receipt & 3.0\% & 3.6\% \\
\midrule
Decision points$^a$ & 19,827 & 3,457 \\
\quad mothers & 2,230 & 396 \\
\quad closed, obs. & 4,225 & 367 \\
\quad closed, none obs. & 8,526 & 2,072 \\
Burden sample$^b$ & 49,256 & 10,789 \\
\bottomrule
\end{tabular}}
\end{table}

\paragraph{Selection and truncation.} Both exports contain only mothers in contact in the export month, so mothers who had already left cannot appear. In the main export, each history is also left-truncated. For the 131 mothers in both exports, the main export contains every row of the replication export inside its own period but none of 1,973 earlier bot rows and 1,106 earlier inbound rows. Retained histories are contiguous and recent, and the selection rule is undocumented. Results therefore describe mothers still on the send list at the end of each export, not all mothers who ever enrolled.

\paragraph{Cohorts.} The window analysis uses all bot messages. The policy evaluation uses bot-initiated free-form messages with at least 7 days of follow-up before the end of the export (decision points). The burden analysis uses delivered bot-initiated messages. The replication export is analysed with the same code and no changes.

\section{Problem Formulation}
\label{sec:formulation}

For bot message $i$ sent at $t_i$, let $\tau_i$ be the mother's most recent message at or before $t_i$, taken from every row with a message from her, and $h_i = t_i - \tau_i$ in hours. We write $W_i$ for the true window state, which a live scheduler knows, and $\tilde W_i = \mathbf{1}\{h_i \le 24\}$ for the state observed in the export, with $\tilde W_i = 0$ when no earlier message is observed. Histories are contiguous, so $\tilde W_i = W_i$ whenever an earlier message is observed; without one, the window may still be open (\appendixref{app:data}).

A scheduler wants to deliver a proactive message at time $t$ with context $x$ and window state $W$. It can send free-form now ($F$), send an approved template now ($T$), wait and send free-form when she next writes within $H$ hours ($D$), or add a reply button to the scheduled template and send the follow-up only if she replies within 24 hours ($T^+$). A message delivered $u$ hours late is worth $e^{-\delta u}$, with $\delta = \ln 2/\ell$ for a half-life $\ell$ of 6 hours (urgent) or 48 hours (routine). A template costs $c$, as a share of a message's value. The platform fixes $P(\text{delivered} \mid F, W = 0) = 0$; we also take $P(\text{delivered} \mid F, W = 1) = 1$, as observed (0 of 35,500 failed). $W$ is known before acting, and $x$ denotes the other features available to a learner, which exclude $h$. Propositions 1 to 3 concern one proactive message with actions $F$, $T$, $D$ or nothing; Proposition 4 concerns a follow-up chained to a scheduled template. Proofs are in \appendixref{app:proofs}.

We measure a policy by its expected value: discounted deliveries minus template costs. In-window free-form messages cost nothing during the study period. Write $p_T(x)$ for the probability that a template sent with the window closed is delivered.

\begin{proposition}[Reach ceiling]\label{prop:ceiling}
A policy that always sends free-form achieves at most $P(W = 1)$. A window-aware policy that sends $F$ when $W = 1$ and, when $W = 0$, the best of $T$, $D$ and sending nothing achieves exactly
\[ P(W = 0)\,\mathbb{E}\big[\max\{p_T(x) - c,\ V_D(x),\ 0\} \mid W = 0\big] \]
more, where $V_D(x)$ is the value of waiting.
\end{proposition}

\begin{proposition}[Known-zero action]\label{prop:known-zero}
A learner that observes $W$ and masks $F$ when $W = 0$ incurs no regret or exploration from that action. If $(x_t, W_t)$ are drawn independently across rounds, a learner whose policy depends on $x$ and its history but not on $W$ has expected regret of at least $gn$ after $n$ rounds against the window-aware oracle. The constant $g$ is positive whenever, on a set of contexts with positive probability, both window states can occur and the best action differs between them.
\end{proposition}

\begin{proposition}[When to wait]\label{prop:wait}
Suppose the mother next writes after a time $R$ that is exponential with rate $\lambda$, and a free-form message sent when she writes is delivered. The value of waiting up to $H$ hours, with no fallback, is
\[ V_D = \mathbb{E}\big[e^{-\delta R}\,\mathbf{1}\{R \le H\}\big] = \frac{\lambda}{\lambda+\delta}\big(1 - e^{-(\lambda+\delta)H}\big), \]
and waiting beats a template if and only if $V_D \ge p_T(x) - c$. For any distribution of $R$, $V_D \le P(R \le H)$, and in the exponential case also $V_D \le \lambda \min(H, 1/\delta)$.
\end{proposition}

\begin{proposition}[Prompting a reply]\label{prop:reply}
Suppose a reply button on the scheduled template draws a reply within 24 hours with probability $q(x)$ per scheduled send (including the chance that the template is not delivered), after a lag $L$, and the follow-up then sent in the open window costs $c_s \ge 0$. Then $T^+$ beats $T$ if and only if
\[ q(x)\big(\mathbb{E}[e^{-\delta L} \mid \text{reply}] - c_s\big) \ge p_T(x) - c, \]
and $q(x) \ge p_T(x) - c$ is a necessary condition.
\end{proposition}

So the window should be encoded, not learned, and neither waiting nor a reply prompt can be worth more than the chance that the mother writes within the deadline (\figureref{fig:logic} in \appendixref{app:proofs}).

\section{Methods}
\label{sec:methods}

We keep three kinds of result apart: \emph{observed} outcomes from the logs, \emph{modelled} counterfactuals for actions not taken, and \emph{simulated} outcomes from a calibrated model, and, as in observational health studies that separate prediction from causal claims \citep{noroozizadeh2025impact}, we make causal claims only where the design supports them (the discontinuity). Full definitions are in \appendixref{app:data} to \appendixref{app:sendtime}.

\subsection{Identifying the loss}
We report failure by format and window with Wilson intervals \citep{wilson1927probable}, and estimate the jump at 24 hours with a local linear probability model \citep{imbens2008regression,lee2010regression,cattaneo2020practical},
\[ Y_i = \alpha + \beta\,\mathbf{1}\{h_i > 24\} + \gamma\,r_i + \eta\, r_i\mathbf{1}\{h_i > 24\} + \varepsilon_i, \]
where $Y_i = 1$ if message $i$ failed, $r_i = h_i - 24$ and $\beta$ is the jump, with bandwidths 6 to 72 hours and standard errors clustered by mother \citep{cameron2015practitioner}. We report conventional intervals across bandwidths rather than bias-corrected intervals \citep{calonico2014robust}; with a jump close to the maximum possible, the choice does not affect the conclusion. We add placebo cut-offs with the same model, the same model for templates, a difference in discontinuities between formats \citep{grembi2016fiscal}, and Fisher's exact test between 18 and 30 hours. Messages with no earlier message observed have no $h_i$ and are excluded; because $h_i \ge 0$, bandwidths above 24 hours widen the window only to the right. Messages are scheduled by flow, not by the time of the mother's last message, which supports continuity of other determinants of failure at 24 hours.

\subsection{Mechanism}
We count failures by flow and examine the timing of the medical-professional offer relative to the mother's last message. Meta error codes are not in the exports. We infer them from format (re-engagement error 131047 for failed free-form messages, undeliverable 131026 for failed templates), and treat the inferred split as descriptive.

\subsection{Return to the chat}
Conversations are runs of the mother's own messages, split at gaps over 12 hours. Kaplan-Meier curves \citep{kaplan1958nonparametric}, censored at the end of the export, give time to the next conversation, and a Cox model \citep{cox1972regression} with a mother-cluster bootstrap relates return to the number of earlier conversations.

\subsection{Offline policy evaluation}
Decision points are bot-initiated free-form messages with 7 days of follow-up. A logistic model, cross-fitted in two folds by mother \citep{chernozhukov2018double}, gives template delivery $\hat p_T(x)$, trained on closed-window templates. Features are whether an earlier message was observed, $\log(1 + \text{earlier messages})$, $\log(1 + h)$ with $h$ capped at 60 days and set to the cap when no earlier message is observed, weekend and evening. We fit it twice: with ``not failed'' as the outcome, an upper bound that counts unconfirmed sends as delivered, and with confirmed delivery, a lower bound. With the window closed, the \emph{logged} policy is what the chatbot did; the \emph{template rule} sends a template; the \emph{wait rule} waits up to 7 days; the \emph{threshold rule} applies \propositionref{prop:wait} with a return rate estimated from a cross-fitted return model; and a \emph{hindsight} bound takes the better of the template's expected value and waiting's realised value. No approved template existed for the main contents affected, so template values are a design counterfactual. We therefore report results by content type and add a \emph{suppress} option for content whose value is doubtful when late. Each of 200 bootstrap replicates resamples mothers in both the decision points and the template training data and refits both models. A replay with linear Thompson sampling \citep{agrawal2013thompson,chapelle2011empirical} compares a learner that masks $F$ when the window is closed, one that sees the window but does not mask, and one that cannot see it.

\paragraph{Identification assumptions.} The modelled template values rest on four assumptions, which we state rather than test. The first is transport: a new approved template for the offer or follow-up would be delivered like existing closed-window templates with the same features, which we check against the observed delivery of the scheduled weekly-guide template sent to the same mothers. The second is positivity: closed-window templates, 45,538 of them in the training data, cover the feature range of the decision points. The third is that waiting does not change return, so under the wait rule a mother writes again when the logs say she did. The fourth is that value does not depend on format, so a message delivered as a template is worth the same as the free-form version. It is the weakest: a template can carry the same buttons, but whether mothers tap them, and whether Meta approves it as utility, is untested, so our outcome is reach, not connection to a clinician.
Importance-weighted and doubly robust estimators \citep{li2011unbiased,dudik2011doubly,jiang2016doubly} are not available without logged propensities, so none of our offline values is design-unbiased. They are model-based estimates, reported with bounds.

\subsection{Simulation and Q-learning}
To let today's action change tomorrow's window, we simulate mothers for 26 weeks. Gaps between conversations follow a two-component exponential mixture fitted to the censored gaps. Each week the chatbot sends the guide as a template and wants to deliver the medical offer. Free-form is delivered only with the window open; templates are delivered at the scheduled guide's observed rate; and a delivered template draws a genuine reply (excluding opt-out taps) at the observed rate, which reopens the window. In the logs, opting out removes a mother from one stream, not the service: of 265 first opt-outs with 30 days of follow-up, 259 (97.7\%) were followed by further proactive messages. The main setting therefore leaves the guide and offer stream untouched by opt-outs; a global model in which each delivered message ends all contact with probability $\rho = 1.69\%$ gives an upper bound on burden.

We compare the logged behaviour, waiting, $T^+$, a hybrid (button, then template after 6 hours), the template rule, the threshold rule and tabular Q-learning \citep{watkins1992qlearning,sutton2018reinforcement}. Q-learning sees time since her last message (five bands) and conversations in the last 28 days (three bands). Its weekly reward is the guide delivered plus the discounted offer value, minus 0.05 per delivered message, with step size $n(s,a)^{-0.6}$ \citep{evendar2003learning}, $\gamma = 0.97$ and exploration decaying from 0.3 to 0.01 over 150,000 training mothers. We validate the simulator against log statistics it was not fitted to, and run 13 single-input stress tests, retraining Q-learning each time. Nothing was tuned to favour any strategy. With weekly decisions and no fatigue state, sending in the main setting has no future cost, so the template rule is optimal there by construction; the test is whether replies that restart the return process, or exit under the global model, change this (\appendixref{app:sim}).

\subsection{Burden and send time}
A learner could beat a fixed rule if sending more messages made mothers more likely to quit, a dynamic the rule ignores. We test whether opting out or reading within 24 hours changes with the messages a mother received in the previous 7 days, comparing each mother with herself (mother, flow and month fixed effects, errors clustered by mother) and, because opting out cuts later volume and biases within-mother estimates downward, with a pooled discrete-time model for her first opt-out. Within batched broadcasts, we compare reading by send time with campaign fixed effects; batch order is undocumented, so this is an association. Models were fitted with statsmodels \citep{seabold2010statsmodels}.

\section{Results}
\label{sec:results}

\subsection{The window explains most of the loss}
\label{sec:res-rq1}

Of 98,607 bot messages, 21,456 (21.8\%) failed. Failures followed the window (\tableref{tab:cells}, \figureref{fig:cliff}). No free-form message failed with the window open, and only one of all failures occurred with it open, a template. Templates failed at 3.3\% when an earlier message was observed and 26.6\% when not, consistent with numbers no longer on WhatsApp rather than the window.

\begin{table}[t]
\floatconts{tab:cells}{\caption{Failure by format and window state, main export (observed). Unconfirmed: status ``sent'' or ``unknown''. None obs.: no earlier message observed.}}{%
\small\setlength{\tabcolsep}{4pt}
\begin{tabular}{@{}llrrr@{}}
\toprule
Format & Window & $n$ & Failed & Unconf. \\
\midrule
Free-form & Open & 35,500 & 0.0\% & 0.1\% \\
 & Closed & 6,267 & 98.6\% & 0.0\% \\
 & None obs. & 8,807 & 90.9\% & 0.0\% \\
Template & Open & 1,690 & 0.1\% & 0.0\% \\
 & Closed & 21,736 & 3.3\% & 4.3\% \\
 & None obs. & 24,607 & 26.6\% & 7.8\% \\
\bottomrule
\end{tabular}}
\end{table}

Between 18 and 30 hours, 0 of 114 free-form messages failed before 24 hours and 72 of 79 after (Fisher's exact test, $p = 2.1 \times 10^{-45}$). The jump at 24 hours is 93.0 percentage points (95\% CI 87.0 to 99.1), and between 89.1 and 95.1 across bandwidths from 6 to 72 hours (\figureref{fig:rd} and \tableref{tab:rd} in \appendixref{app:data}). Templates show none (1.0 point, CI $-0.7$ to 2.7); the difference in discontinuities is 92.0 points (CI 85.6 to 98.4). With the same local linear model, no placebo cut-off shows a positive jump; every interval includes zero, the one at 48 hours only just ($-14.1$, CI $-28.1$ to 0.0), and template placebos are all null (\appendixref{app:data}).

\paragraph{Replication.} In the replication export, 0 of 9,546 free-form messages failed with the window open, and between 18 and 30 hours 0 of 19 failed before 24 hours against 18 of 18 after (\appendixref{app:replication}).

\subsection{Two flows produce two thirds of the loss}
\label{sec:res-rq2}

The error follow-up (9,525 failures, 44.4\% of all) and the medical-professional offer (4,654, 21.7\%) account for 66.1\% of failures (\figureref{fig:flows} in \appendixref{app:data}). Both send free-form content to mothers whose window has closed.

\paragraph{The offer fails by design.} It is chained a second after the weekly guide, a template, and a template does not open the window. Of 4,869 offers, 95.6\% failed: all 2,080 sent with the window closed and 2,574 of 2,583 with no earlier message observed, but none of 206 sent with it open (\figureref{fig:handoff} in \appendixref{app:data}). Where an earlier message was observed, the median time since it was 181 hours. In practice, an offer to connect a pregnant woman with a clinician reached almost no one.

\paragraph{Mothers return slowly.} Across 9,278 conversations, the median time to the next one was 51 days. After a first conversation, 13.6\% of mothers wrote again within 7 days and 36.3\% within 30 days (\figureref{fig:km} in \appendixref{app:decision}). Waiting for the window to reopen therefore rarely works.

\subsection{Scheduling-policy evaluation}
\label{sec:res-rq3}

The evaluation uses 19,827 decisions from 2,230 mothers; 12,751 were sent with the window closed or no earlier message observed (4,225 and 8,526): 64.4\% error follow-ups, 30.8\% medical offers and 4.8\% other content (\tableref{tab:content}). Template values are modelled.

\begin{table*}[t]
\floatconts{tab:content}{\caption{Closed-window decisions by content, main export, per 100 (95\% intervals from 200 mother-cluster bootstrap replicates that refit the delivery models). Logged values are observed; wait values replay the logs under the assumption that waiting does not change return; template values are modelled.}}{%
\small\setlength{\tabcolsep}{4pt}
\begin{tabular}{@{}lrrrrr@{}}
\toprule
Content & $n$ & Logged & Template, confirmed & Template, not failed & Wait 7 days \\
\midrule
Medical offer & 3,933 & 0.3 (0.2 to 0.5) & 81.5 (79.8 to 83.0) & 86.8 (85.7 to 87.9) & 16.6 (14.8 to 18.1) \\
Error follow-up & 8,209 & 1.5 (1.2 to 2.0) & 80.9 (79.3 to 82.5) & 86.4 (85.2 to 87.5) & 17.2 (15.5 to 18.6) \\
Other content & 609 & 99.5 (98.8 to 100) & 67.5 (65.2 to 69.6) & 75.0 (73.1 to 77.2) & 88.8 (84.1 to 92.8) \\
\bottomrule
\end{tabular}}
\end{table*}

\begin{figure}[t]
\floatconts{fig:offer}{\caption{Options for the medical-professional offer per 100 closed-window decisions ($n$ = 3,933), main export, with 95\% intervals (right). Logged: observed. Wait: offline replay of observed returns. Reply proxy: any message within 24 hours (no button was deployed). Template: hypothetical approved-template counterfactual (modelled, open markers), bounded by confirmed and not-failed delivery. Deliveries, not clinician contacts.}}{\includegraphics[width=\columnwidth]{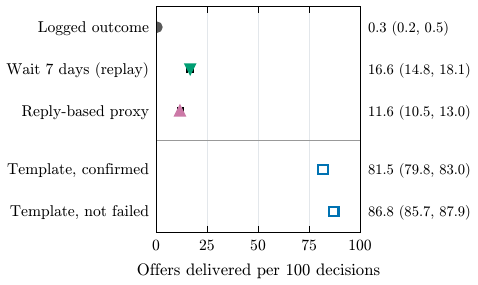}}
\end{figure}

\paragraph{Medical offer.} As a template, the offer would reach an estimated 81.5 to 86.8 in 100 mothers, against 0.3 as logged, 11.6 (CI 10.5 to 13.0) for a reply button and 16.6 for waiting (\figureref{fig:offer}). The reply-button value counts any message from the mother within 24 hours of the offer; the guide had no button, so this understates a button (\appendixref{app:decision}). Applying the scheduled-guide template's observed delivery instead of the model gives 81.5 to 88.2, so the result does not depend on the model.

\paragraph{Error follow-up.} No template existed for it, it would probably not qualify as a utility template, and its value days later is doubtful. The realistic comparator is suppression, so we do not count follow-up deliveries as a gain for the template rule.

\paragraph{Other content.} 603 of 609 had no earlier message observed and were delivered as logged. They come from onboarding and menu flows in which the mother's opening message is missing from the export, the clearest sign that the ``no earlier message observed'' group contains open windows.

\paragraph{Pooled results.} The template rule delivers 86.0 per 100 closed-window decisions (80.4 confirmed only), against 5.8 logged and 20.4 for waiting. The threshold rule never waits, so it equals the template rule; the hindsight bound reaches 87.7. Waiting overtakes the template only when a template costs 75\% to 85\% of a message's value, depending on the half-life (\figureref{fig:map}). Cross-fitted delivery predictions are within 7.5 points of observed delivery in every decile (\tableref{tab:calib}). On the 4,225 decisions with an earlier message observed, where the window state is exact, the template reaches 97.3 to 99.0 per 100, waiting 28.8 and the logged policy 0.1 (\tableref{tab:restricted}).

\paragraph{Learning on the logs.} In the replay, a learner that masks $F$ has regret 1.4 per 100 decisions, one that sees the window without masking 1.5, a window-blind learner 2.3 and the logged behaviour 49.7 (\figureref{fig:bandit}). The blind learner almost never sends free-form into a closed window (0.3\%); its extra regret comes from sending templates when the window is open (50\% of those decisions), as \propositionref{prop:known-zero} predicts.

\subsection{Simulation: learning converges to the rule}
\label{sec:res-rq4}

Under the logged behaviour, the simulator reproduces outcomes it was not fitted to: offer failure is 98.7\% simulated against 99.4\% observed. It slightly understates open windows (1.5\% against 3.1\% of offers) and overstates later returns (43.2\% against 37.8\% within 30 days), both of which favour waiting (\figureref{fig:val}).

\begin{figure*}[t]
\floatconts{fig:sim}{\caption{Simulated strategies over 26 weeks: (a) offers delivered and (b) discounted value per 100 offers due, main setting (no mother leaves); (c) share who leave by week 26 under the global opt-out model, an upper bound, not observed attrition. Means with 95\% intervals over 8 seeds of 3,000 mothers. The template rule, threshold rule and Q-learning reach the same policy.}}{\includegraphics[width=\textwidth]{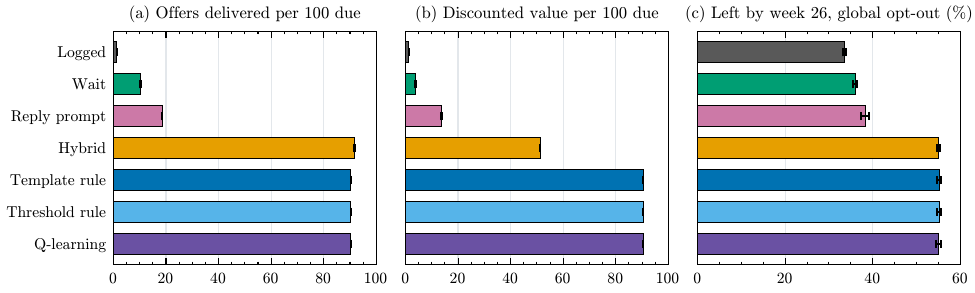}}
\end{figure*}

The template rule delivers 90.3 offers per 100 due, against 1.3 for the logged behaviour, 10.2 for waiting and 18.5 for a reply button (\figureref{fig:sim}, \tableref{tab:strategies}). The hybrid delivers 91.7 but loses about half their value to delay (discounted value 51.2). After about 10,000 training mothers, Q-learning sends free-form when the window is open and a template in every other state. Its reward then equals the template rule's (44.6 per mother) to within simulation noise, and it never finds a better policy. Across 13 single-input stress tests, the template rule, the threshold rule and Q-learning tie (largest gap 0.05 reward units) and beat every other strategy; the closest challenger is the hybrid for routine content, 1.4 units behind (\figureref{fig:stress}). Under the global opt-out bound, delivering the offer raises the share who leave by week 26 from 33.5\% to 55.1\%, and the template rule still has the highest reward. Allowed to suppress the offer, Q-learning matches the template rule until exit per message reaches eight times the bound (\tableref{tab:suppress}). Each extra offer costs about one delivered message, US\$0.010 at the rate used here.

\subsection{Burden and send time}
\label{sec:res-burden}

Across 49,256 delivered messages with 785 opt-outs, within-mother estimates were negative: per log unit of templates in the previous week, opting out within 24 hours changed by $-2.3$ points ($-1.6$ per doubling of one plus the count), controlling for recency (\tableref{tab:burden}). These are biased downward, because opting out cuts later volume; a pooled model of the first opt-out shows no association (odds ratio 0.94 per log unit, CI 0.75 to 1.17). We find no rise, not a fall. Reading within 24 hours fell by 2.2 points per doubling, from about 60\%. Within 17 batched broadcasts, messages sent after 22:00 were read within 6 hours 23.0 points less often than morning sends (CI $-30.9$ to $-15.2$), and by 19.5 to 23.8 points when each campaign is dropped; 193 of the 266 night messages come from one campaign. Prior reading and contact were balanced (joint $p = 0.86$ and $0.36$). Reading eventually did not differ, so late sends mainly wait until morning (\figureref{fig:time}).

\section{Discussion}
\label{sec:discussion}

\subsection{The rule is the problem, and the fix}
The window explains most failed messages, with a jump of about 93 points at 24 hours, replicated in a second export. One of the two flows behind the loss is a clinician referral that, as built, could reach only mothers who had written in the previous day. Once a suitable template exists, a template reaches 81 to 87 in 100 mothers whose window has closed, while waiting or prompting a reply reaches 12 to 17. \propositionref{prop:wait} explains why: waiting can deliver no more often than the mother writes within the deadline, which happened for about 20 in 100 closed-window decisions.

\subsection{Implications for machine learning in health}
A Q-learning scheduler with the same choices learns the template rule and stops: the platform rule makes the most important part of the decision known in advance. When a rule fixes the value of an action and the rule is observable, the action needs neither exploration nor estimation, and a learner that cannot see the rule pays a constant cost in every round (\propositionref{prop:known-zero}).

A learner can beat a fixed rule only where the problem has structure the rule ignores. We tested the most plausible candidate, burden that builds with volume, and found none. Learning may still help with other decisions, such as which content to send or at what hour. We recommend removing every action whose value a platform fixes before learning, then asking what structure is left. This follows guidelines that ask what a learned policy does differently from current practice \citep{gottesman2019guidelines}.

\subsection{Practical deployment considerations}
What follows are design recommendations, not tested interventions. The scheduler should check the window before every proactive free-form message and send a template, or nothing, when it is closed. The medical offer needs its own utility template; Meta decides the category and may recategorise templates \citep{hughey2026observability}, so approval is a risk to plan for. The error follow-up can simply be suppressed when the window is closed. Operators can monitor failures by flow and hours since the mother's last message. Costs are small: about US\$0.47 per mother over six months in simulation, against US\$0.24 as logged, at US\$0.0101 per message and counting in-window service messages as charged, as under pricing from October 2026 \citep{meta_pricing}. Billed as marketing (an indicative US\$0.0516; \citealp{ominiflow2026nigeria}), the offer would raise this to about US\$1.43 and face per-user limits; the non-profit exemption for service messages \citep{meta_pricing} barely changes it.

\subsection{Limitations}
Both exports contain only mothers still in contact, with left-truncated histories selected by an undocumented rule; for shared mothers, 12.2\% of messages labelled ``no earlier message observed'' went to mothers who had written. Error codes are inferred, but the window effect is identified from timing. Template results assume a new template delivered like existing ones, and 3\% of messages lack a receipt, so we report bounds. The nearly deterministic logged policy ruled out importance weighting. The simulator is calibrated to the same logs and gives learning little to exploit, so it shows that the obvious dynamics do not change the answer, not that no richer dynamics could. Whether a template offer leads mothers to a clinician as often as a free-form one would is untested. All outcomes are deliveries, reads or replies, and nothing here shows that clinical outcomes improved. Finally, the evidence comes from one service in one country and one language.

\subsection{Ethical considerations}
The loss falls unevenly. Mothers who rarely write, whose phones are off or whose numbers no longer work are the least likely to receive proactive messages, and they may most need a referral; a template rule helps the first group only, and 26.6\% of templates to mothers with no earlier message observed still failed. More messages could also add burden if opting out ended all contact. The analysis used de-identified logs (salted hashes, no names or message text, aggregates only), and no mother was contacted for this study.

\section{Conclusion}

In a Nigerian maternal health chatbot, WhatsApp's 24-hour window explains most failed messages; approved templates would recover most of the loss, and reinforcement learning rediscovers the simple rule. The first step is not a smarter algorithm but checking the window.

\section*{Data and Code Availability}
The data are conversation logs from a WhatsApp maternal health chatbot operated by a non-profit organisation in Nigeria. They contain phone numbers, names and the text of health questions, and are not publicly available; access depends on the operator's data-sharing terms. Analysis code, tests, aggregate result tables and figures are provided as a code archive under the MIT licence, with a README describing the pipeline, a reproducibility manifest (software versions, random seeds, sample counts and checksums) and unit tests that run without the private data.

\section*{Institutional Review Board (IRB)}
The research does not require IRB approval. It is a retrospective secondary analysis of de-identified logs collected during routine operation of the service; no mother was contacted and no message was sent for this study.

\acks{We thank HelpMum Africa for developing and operating the chatbot service and supporting this research.}

\bibliography{refs}

\clearpage
\appendix

\section*{Outline of the Appendix}
\label{app:outline}

The appendix gives the proofs, the variable definitions, the details of each model and the robustness checks behind the main text. \tableref{tab:outline} lists what each part contains. Observed, modelled and simulated results are labelled as in the main text, and every table and figure caption states its data source.

\begin{table}[h]
\floatconts{tab:outline}{\caption{Contents of the appendix, with the tables and figures in each part.}}{%
\small
\begin{tabular}{@{}lp{0.85\linewidth}@{}}
\toprule
 & Contents \\
\midrule
\ref{app:proofs} & Proofs of Propositions 1 to 4; the hindsight bound; the scheduling logic (\figureref{fig:logic}) \\
\ref{app:data} & Variables, observed and true window state, truncation check, error codes, discontinuity and placebos (\tableref{tab:rd}, \tableref{tab:placebo}, \figureref{fig:handoff}, \figureref{fig:rd}, \figureref{fig:flows}) \\
\ref{app:decision} & Decision models and feature coding, sample reconciliation, transport check, calibration, restricted samples, bandit replay (\tableref{tab:calib}, \tableref{tab:restricted}, \tableref{tab:replay}, \figureref{fig:km}, \figureref{fig:map}, \figureref{fig:bandit}) \\
\ref{app:sim} & Simulator inputs and dynamics, Q-learning details, strategies and costs, suppressing the offer, validation, stress tests (\tableref{tab:strategies}, \tableref{tab:suppress}, \figureref{fig:val}, \figureref{fig:cost}, \figureref{fig:stress}) \\
\ref{app:burden} & Burden definitions, within-mother and pooled models (\tableref{tab:burden}) \\
\ref{app:sendtime} & Send time within batched broadcasts, clustering (\figureref{fig:time}) \\
\ref{app:replication} & Replication on the second export (\tableref{tab:replication}) \\
\ref{app:repro} & Software, seeds and tests \\
\bottomrule
\end{tabular}}
\end{table}

\section{Propositions and Proofs}
\label{app:proofs}

\begin{figure*}[htbp]
\floatconts{fig:logic}{\caption{Window-aware scheduling logic implied by Propositions 1, 3 and 4, for one proactive message. $D$ and $T^+$ are chained actions: under $D$ the free-form message is sent only if she writes within $H$, and under $T^+$ the follow-up is sent only if she replies to the scheduled template within 24 hours, so its value is $q(x)$ times the value of the follow-up net of its cost $c_s$ (\propositionref{prop:reply}). A proposed design: it was not deployed or tested as an intervention in this study.}}{\includegraphics[width=0.85\textwidth]{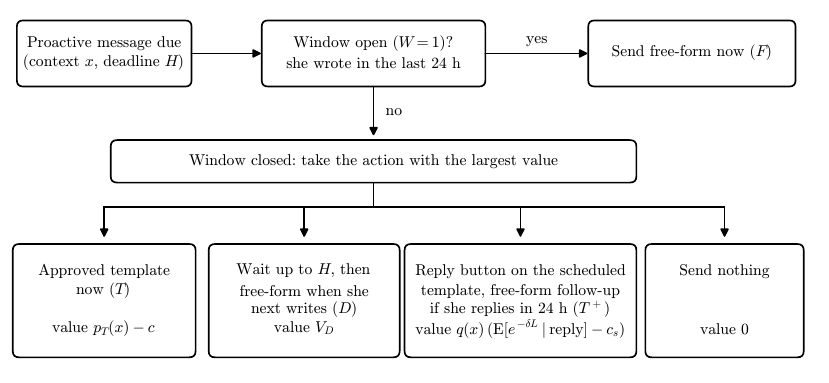}}
\end{figure*}

Throughout, $\mu_a(x, W)$ is the expected value of action $a$ in context $x$ and window state $W$, and $\mu^*(x, W) = \max_a \mu_a(x, W)$. With the window closed, $\mu_F(x, 0) = 0$, $\mu_T(x, 0) = p_T(x) - c$, $\mu_D(x, 0) = V_D(x)$ and sending nothing has value 0.

\paragraph{Proof of \propositionref{prop:ceiling}.} A policy that always sends $F$ has value $P(W{=}1)\,P(\text{delivered} \mid F, W{=}1) + P(W{=}0) \cdot 0 \le P(W{=}1)$, because $P(\text{delivered} \mid F, W{=}0) = 0$ and in-window messages cost nothing. The window-aware policy earns the same when $W = 1$. When $W = 0$ it earns $\max\{p_T(x) - c, V_D(x), 0\}$ given $x$, against 0 for the always-$F$ policy. Taking expectations over $x$ given $W = 0$ and weighting by $P(W{=}0)$ gives the stated difference, which is never negative. $\square$

\paragraph{Proof of \propositionref{prop:known-zero}.} The regret in round $t$ is $r_t = \mu^*(x_t, W_t) - \mu_{a_t}(x_t, W_t) \ge 0$. Let $\mathcal{H}_{t-1}$ be the history of earlier contexts, actions and rewards, and $\pi_t(a \mid x_t, W_t, \mathcal{H}_{t-1})$ the learner's policy. Choosing $F$ when $W_t = 0$ contributes $\mu^*(x_t, 0)$, so
\[ \mathbb{E}[r_t] \ge \mathbb{E}\big[\mathbf{1}\{W_t{=}0\}\,\pi_t(F \mid x_t, W_t, \mathcal{H}_{t-1})\,\mu^*(x_t, 0)\big]. \]
A learner that observes $W$ and masks $F$ has $\pi_t(F \mid x_t, 0, \mathcal{H}_{t-1}) = 0$, so this term vanishes, and because $\mu_F(x, 0) = 0$ is known it never needs to try $F$ to learn it. Now take a learner whose policy $\pi_t(a \mid x_t, \mathcal{H}_{t-1})$ does not depend on $W_t$, with $(x_t, W_t)$ drawn independently across rounds from a fixed distribution, so that $W_t$ is independent of $\mathcal{H}_{t-1}$ given $x_t$. Conditional on $x_t$ and $\mathcal{H}_{t-1}$, it chooses $a$ with a fixed probability in both window states, so its expected value is $\sum_a \pi_t(a \mid x_t, \mathcal{H}_{t-1})\,\mathbb{E}[\mu_a(x_t, W) \mid x_t] \le \max_a \mathbb{E}[\mu_a(x_t, W) \mid x_t]$. Its expected regret in every round is therefore at least
\[ g = \mathbb{E}_x\Big[\mathbb{E}[\mu^*(x, W) \mid x] - \max_a \mathbb{E}[\mu_a(x, W) \mid x]\Big] \ge 0, \]
and at least $gn$ after $n$ rounds, whatever it learns. Let $S$ be the set of contexts with $0 < P(W{=}1 \mid x) < 1$ in which no action is best in both window states. For $x \in S$, every action falls short of $\mu^*(x, w)$ in some state $w$ of positive probability, so $\mathbb{E}[\mu_a(x, W) \mid x] < \mathbb{E}[\mu^*(x, W) \mid x]$ for each of the finitely many actions, and the bracket is positive. Hence $P(S) > 0$ implies $g > 0$. If $x$ included $h$, then $P(W{=}1 \mid x) \in \{0, 1\}$, $S$ would be empty and $g = 0$; this is why $x$ excludes $h$. In the replay, the window-aware learners see $\log(1 + h)$ and the window indicator, while the window-blind learner sees neither (\appendixref{app:decision}). $\square$

In our setting $S$ is large: with the window open the best action is $F$, which is free and delivered with probability 1 by assumption, and with it closed $F$ is worth nothing.

\paragraph{Proof of \propositionref{prop:wait}.} A message sent when the mother writes at time $R \le H$ is delivered $R$ hours late and is worth $e^{-\delta R}$; if $R > H$, nothing is delivered. Hence $V_D = \mathbb{E}[e^{-\delta R}\mathbf{1}\{R \le H\}]$. Because $e^{-\delta R} \le 1$, $V_D \le P(R \le H)$ for any distribution of $R$. For $R$ exponential with rate $\lambda$,
\[ V_D = \int_0^H \lambda e^{-\lambda r} e^{-\delta r}\,dr = \frac{\lambda}{\lambda + \delta}\big(1 - e^{-(\lambda+\delta)H}\big). \]
Dropping $e^{-\lambda r} \le 1$ inside the integral gives $V_D \le (\lambda/\delta)(1 - e^{-\delta H}) \le \lambda \min(H, 1/\delta)$, since $1 - e^{-y} \le \min(y, 1)$. Waiting has no template cost, so it beats a template if and only if $V_D \ge p_T(x) - c$. $V_D$ increases with $H$ and $\lambda$ (a larger rate makes $R$ stochastically smaller, and the integrand is decreasing in $R$) and decreases with $\delta$. $\square$

\paragraph{Remark on estimation.} The threshold rule uses $\hat\lambda = -\ln(1 - \hat p_R)/H$, where $\hat p_R$ is the predicted probability of a return within $H$, so the exponential model matches $P(R \le H)$. If the return hazard falls over time, as in the mixture fitted in \appendixref{app:sim}, returns within $H$ come earlier than this constant rate implies and the formula understates $V_D$. The bound $V_D \le P(R \le H)$ holds for any hazard, so, with a calibrated $\hat p_R$, the template is preferred whenever $p_T(x) - c \ge \hat p_R(x)$, whatever the shape of the hazard. In the evaluation, the value of the wait rule itself is computed from observed returns, not from this formula.

\paragraph{Proof of \propositionref{prop:reply}.} The scheduled template is sent under both $T$ and $T^+$, so its value and cost cancel. Under $T^+$, the follow-up is sent only after a reply, when the window is open and a free-form message is delivered; its value is $\mathbb{E}[e^{-\delta L} \mid \text{reply}] - c_s$ with probability $q(x)$, and zero otherwise. Because $q(x)$ is per scheduled send, it equals the template's delivery probability times the reply rate among delivered templates; the simulator uses these two factors separately. Under $T$, the follow-up is a second template with value $p_T(x) - c$. Comparing the two gives the condition. Because $\mathbb{E}[e^{-\delta L} \mid \text{reply}] - c_s \le 1$, the value of $T^+$ is at most $q(x)$, so $q(x) \ge p_T(x) - c$ is necessary. $\square$

\paragraph{Hindsight bound.} For each decision, the hindsight policy takes the larger of the template's expected value and waiting's realised value. It bounds any policy that knows when each mother will return but not whether a template will be delivered.

\section{Data and Variables}
\label{app:data}

\paragraph{Variables.} $Y_i = 1$ if the final status is ``failed''; $D_i = 1$ if ``delivered'' or ``read''; ``sent'' and ``unknown'' are unconfirmed. Times are in West Africa Time. Identifiers are salted SHA-256 hashes of phone numbers; names and message text are excluded from analysis tables.

\paragraph{Window definition.} The window is computed from every row containing a message from the mother, including 1,398 inbound rows with no bot reply. Computing it from bot rows only changed the state of 30 of 98,607 messages and no result.

\paragraph{Observed and true window state.} A live scheduler knows the true state $W_i$ from its own records. The export gives $\tilde W_i$. When an earlier message is observed, histories are contiguous, so $\tilde W_i = W_i$. When none is observed, $\tilde W_i = 0$ but the window may be open: in the truncation check below, 33 of 551 such messages (6.0\%) had an open window, and onboarding and menu flows (the ``other content'' decisions) almost certainly did. Results that condition on an earlier message being observed (the discontinuity, \figureref{fig:cliff}, and \tableref{tab:restricted}) are free of this error.

\paragraph{Truncation check.} For the 131 shared mothers, 67 of 551 messages (12.2\%) labelled ``no earlier message observed'' in the main export went to mothers who had written according to the longer replication history, and 33 had an open window. Results for mothers with an earlier message observed are unaffected because histories are contiguous.

\paragraph{Late deliveries.} The 7 free-form messages delivered between 24 and 30 hours most likely followed replies the export does not record, such as some button taps.

\paragraph{Error codes.} Failed free-form messages are assigned 131047 (re-engagement) and failed templates 131026 (undeliverable). This is an inference from message format, not a recorded code; the operator's own error summary was computed from the same export, so agreement with it is a consistency check, not verification. Because codes follow format, the resulting split of all 21,456 failures, 66.1\% and 33.9\%, restates the split of failures by format in \tableref{tab:cells}; all medical-offer failures are free-form and so carry 131047. The 66.1\% coincides with, but is distinct from, the share of failures from the two flows in \sectionref{sec:res-rq2}.

\paragraph{Discontinuity.} \tableref{tab:rd} gives the jump at each bandwidth. The estimation window is $[\max(0, 24 - b), 24 + b]$ for bandwidth $b$, so for $b \ge 24$ the left side is always 0 to 24 hours and only the right side widens. Messages with no earlier message observed have no $h_i$ and are excluded. Placebo cut-offs at 12, 36, 48, 72 and 120 hours use the same local linear model with a 12-hour bandwidth below 48 hours and 24 hours otherwise (\tableref{tab:placebo}); an earlier version used mean differences, which can mix a slope with a jump: at 72 hours the mean difference was 5.3 points (CI 0.4 to 10.2) and the local linear estimate is $-4.4$ (CI $-16.2$ to 7.4). The difference in discontinuities interacts the jump with format within the 24-hour bandwidth (0 to 48 hours). \figureref{fig:flows} shows which flows produce the failures.

\begin{table}[h]
\floatconts{tab:rd}{\caption{Robustness of the discontinuity to bandwidth. Local linear estimates of the jump in the share of free-form messages that failed at 24 hours after the mother's last message, main export (observed). Percentage points with 95\% intervals clustered by mother; the main text reports the 24-hour bandwidth. Upper limits above 100 arise because the linear probability model's symmetric intervals are not bounded; no message failed just before 24 hours.}}{%
\small
\begin{tabular}{@{}rrr@{}}
\toprule
Bandwidth (h) & Jump & 95\% CI \\
\midrule
6 & 89.8 & 75.9 to 103.7 \\
12 & 89.1 & 80.2 to 98.0 \\
18 & 91.5 & 84.2 to 98.9 \\
24 & 93.0 & 87.0 to 99.1 \\
48 & 95.1 & 91.3 to 99.0 \\
72 & 93.9 & 90.8 to 96.9 \\
\bottomrule
\end{tabular}}
\end{table}

\begin{table}[h]
\floatconts{tab:placebo}{\caption{Placebo cut-offs, main export (observed). Local linear jump in failure at each false cut-off, percentage points, 95\% intervals clustered by mother; bandwidth 12 hours below 48 hours, 24 hours otherwise. At 12 hours no free-form message failed on either side.}}{%
\small\setlength{\tabcolsep}{3pt}
\begin{tabular}{@{}rrrrr@{}}
\toprule
 & \multicolumn{2}{c}{Free-form} & \multicolumn{2}{c}{Templates} \\
Cut-off (h) & Jump & 95\% CI & Jump & 95\% CI \\
\midrule
12 & 0.0 & no failures & 0.0 & 0.0 to 0.1 \\
36 & $-$4.3 & $-$11.6 to 3.0 & 1.5 & $-$3.1 to 6.2 \\
48 & $-$14.1 & $-$28.1 to 0.0 & 0.0 & $-$3.2 to 3.1 \\
72 & $-$4.4 & $-$16.2 to 7.4 & 2.5 & $-$1.9 to 7.0 \\
120 & 4.1 & $-$7.3 to 15.5 & 2.7 & $-$1.9 to 7.2 \\
\bottomrule
\end{tabular}}
\end{table}

\begin{figure*}[htbp]
\floatconts{fig:handoff}{\caption{The medical-professional offer, main export (observed). (a) Histogram of hours since the mother's last message (log scale) for the 2,286 offers with an earlier message observed. (b) Failure rate by window state, with Wilson 95\% intervals and the number of offers. None obs.: no earlier message from the mother in the export, an unobserved history rather than proof that the window was closed.}}{\includegraphics[width=0.85\textwidth]{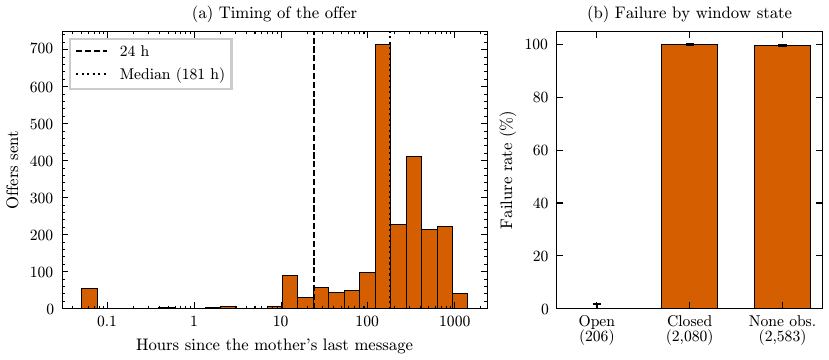}}
\end{figure*}

\begin{figure*}[htbp]
\floatconts{fig:rd}{\caption{Robustness of the jump in failure at 24 hours, main export (observed). (a) Local linear estimates by bandwidth. (b) The true cut-off (24*) and placebo cut-offs with the same model, bandwidth 12 hours below 48 hours and 24 hours otherwise. Percentage points with 95\% intervals clustered by mother. At 12 hours no free-form message failed on either side, so that estimate has no interval.}}{\includegraphics[width=0.8\textwidth]{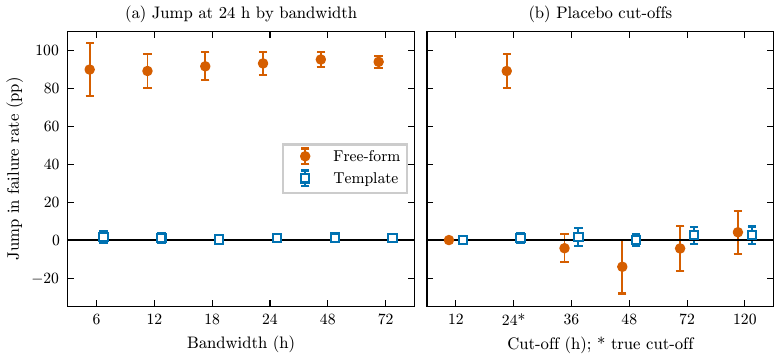}}
\end{figure*}

\begin{figure*}[htbp]
\floatconts{fig:flows}{\caption{Failed messages by flow, main export (observed): the ten flows with the most failures, labelled with failed and sent messages and the percentage that failed. The error follow-up and the medical-professional offer produce 66.1\% of the 21,456 failures.}}{\includegraphics[width=0.85\textwidth]{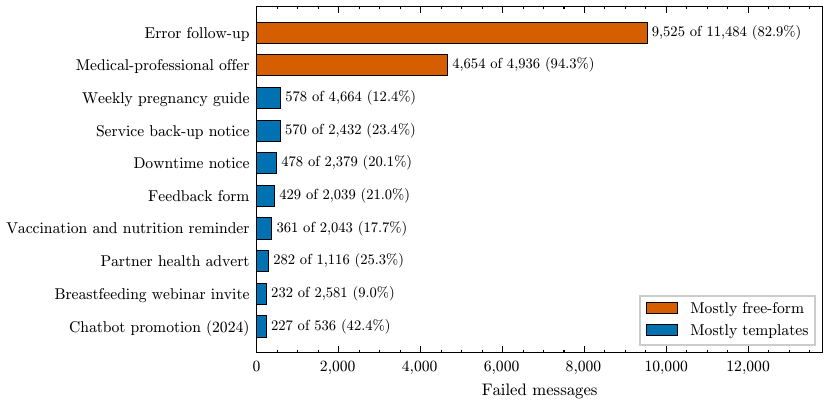}}
\end{figure*}

\section{Decision Models and Offline Evaluation}
\label{app:decision}

\paragraph{Models.} Cross-fitting uses two folds with mothers ordered by first appearance (seed 42). The features are an indicator that no earlier message was observed, $\log(1 + \text{earlier messages})$, which is zero when there are none, $\log(1 + h)$ with $h$ capped at 60 days and set to the cap when no earlier message is observed (so the indicator acts as a separate intercept for that group), weekend and evening (17:00 to 23:59). The template model is trained on the 45,538 bot-initiated templates sent with the window closed or no earlier message observed. Return within 7 days is modelled by logistic regression on the same features, trained on the closed-window decisions, and converted to a rate $\hat\lambda = -\ln(1 - \hat p_R)/H$. Cross-fitted AUCs are 0.78 (template delivery) and 0.69 (return). \tableref{tab:calib} shows calibration. \figureref{fig:km} shows time to the next conversation; mothers with more earlier conversations returned faster (hazard ratio 1.25 per log unit, 95\% CI 1.18 to 1.31; Cox model with a mother-cluster bootstrap).

\paragraph{The medical offer: samples.} The flow contains 4,936 bot messages: the 4,869 interactive offers analysed in \sectionref{sec:res-rq2}, 63 text messages and 4 templates, none of which failed. The 3,933 closed-window decisions are the bot-initiated free-form messages in the flow with 7 days of follow-up: 3,930 interactive offers, of which 9 did not fail, and 3 text messages, all delivered; the logged value of 0.3 per 100 counts all 12. Interactive offers alone give 0.2 (CI 0.1 to 0.4), and the other values change by at most 0.1 (\tableref{tab:restricted}). The simulator's validation targets (\figureref{fig:val}) use the 1,963 offers sent directly after the scheduled weekly guide, the chain it models, which is why its failure rates (99.4\% overall, 96.9\% with an earlier message observed) differ from those of all offers.

\paragraph{Reply button and transport check.} No reply button was ever attached to the guide before the offer, so the reply-button value is the observed share of offers followed by any message from the mother within 24 hours. Templates in other flows that did carry buttons were answered more often (17.9\% within 24 hours, excluding opt-out taps), which is the rate the simulator uses; the offline value of 11.6 is therefore conservative. The transport check replaces the model with the observed delivery of the scheduled weekly-guide template, 90.5\% (87.6\% confirmed) when an earlier message was observed and 86.9\% (78.0\% confirmed) when not, weighted by the 36.1\% and 63.9\% of offer decisions in each group. Refitting inside the bootstrap widens the pooled template interval (85.5 to 86.5 without refitting; 84.9 to 87.1 with it). \tableref{tab:restricted} repeats the evaluation on restricted samples. \figureref{fig:offer} compares the options for the medical offer, \figureref{fig:map} shows when waiting would beat a template and how this depends on template cost and half-life (grid $c \in [0, 0.95]$, $\ell \in \{6, 24, 48, 168\}$ hours), and \figureref{fig:bandit} shows the bandit replay.

\begin{table}[h]
\floatconts{tab:calib}{\caption{Calibration of the cross-fitted template delivery model, main export. Closed-window templates grouped into deciles of predicted delivery (about 4,554 per decile); predicted and observed ``not failed'' delivery, and observed confirmed delivery, in percent.}}{%
\small
\begin{tabular}{@{}rrrr@{}}
\toprule
Decile & Predicted & Observed & Confirmed \\
\midrule
1 & 59 & 62 & 52 \\
2 & 68 & 60 & 52 \\
3 & 75 & 76 & 69 \\
4 & 79 & 83 & 76 \\
5 & 81 & 78 & 70 \\
6 & 89 & 93 & 86 \\
7 & 95 & 95 & 89 \\
8 & 97 & 96 & 91 \\
9 & 99 & 99 & 96 \\
10 & 99 & 99 & 98 \\
\bottomrule
\end{tabular}}
\end{table}

\begin{table*}[h]
\floatconts{tab:restricted}{\caption{Closed-window evaluation on restricted samples, main export, per 100 decisions with 95\% intervals from 500 mother-cluster bootstrap replicates (delivery models not refitted). Logged values are observed; wait values replay the logs (delivered within 7 days by waiting, assuming waiting does not change return); reply values are a proxy (any message from the mother within 24 hours); template values are modelled (lower: confirmed delivery; upper: not failed). With an earlier message observed, the window state is exact.}}{%
\small\setlength{\tabcolsep}{3pt}
\begin{tabular}{@{}lrrrrrr@{}}
\toprule
Sample & $n$ & Logged & Template lower & Template upper & Wait & Reply \\
\midrule
All closed or none obs. (paper) & 12,751 & 5.8 & 80.4 & 86.0 & 20.4 & 15.2 \\
\quad 95\% CI & & 5.2 to 6.5 & 79.8 to 81.1 & 85.5 to 86.5 & 18.9 to 22.0 & 13.9 to 16.5 \\
Earlier message observed & 4,225 & 0.1 & 97.3 & 99.0 & 28.8 & 20.6 \\
\quad 95\% CI & & 0.0 to 0.1 & 97.2 to 97.4 & 98.9 to 99.0 & 26.4 to 31.4 & 18.4 to 22.7 \\
Offer and error follow-up & 12,142 & 1.1 & 81.1 & 86.5 & 17.0 & 11.6 \\
\quad 95\% CI & & 0.8 to 1.4 & 80.4 to 81.8 & 86.0 to 87.1 & 15.5 to 18.5 & 10.4 to 12.9 \\
Offer, interactive only & 3,930 & 0.2 & 81.5 & 86.8 & 16.5 & 11.5 \\
\quad 95\% CI & & 0.1 to 0.4 & 80.8 to 82.2 & 86.3 to 87.4 & 15.3 to 18.1 & 10.4 to 12.8 \\
Offer, interactive, earlier msg. & 1,421 & 0.0 & 97.3 & 99.0 & 28.7 & 21.3 \\
\quad 95\% CI & & 0.0 to 0.0 & 97.2 to 97.4 & 98.9 to 99.0 & 26.2 to 31.6 & 19.0 to 23.9 \\
\bottomrule
\end{tabular}}
\end{table*}

\paragraph{Bandit replay.} Decision points are replayed in time order. Each learner chooses among $F$, $T$ and $D$. The reward of $F$ is 1 with the window open and 0 with it closed; of $T$, a Bernoulli draw with probability $\hat p_T(x)$ minus $c = 0.05$, so choosing $T$ with the window open costs at least 0.05; of $D$, the observed discounted value of waiting. Regret uses the expected rewards. Each action has a linear model with prior mean zero, prior precision equal to the identity and noise variance 0.25. The window-aware learners see an intercept, the no-earlier-message indicator, $\log(1 + \text{earlier messages})$, $\log(1 + h)$, weekend, evening and the window indicator; the window-blind learner sees the same without $\log(1 + h)$ and the window indicator. Results are means over 30 runs (seeds 0 to 29). \tableref{tab:replay} repeats the replay without the 609 closed-window ``other content'' decisions, whose window was probably open, and on decisions with an earlier message observed. Both restrictions lower the regret of the window-aware learners, and the ordering is unchanged: masked, then unmasked, then blind, with the logged behaviour far behind.

\begin{table*}[h]
\floatconts{tab:replay}{\caption{Regret per 100 decisions in the bandit replay on restricted samples, main export (modelled rewards; means over 30 runs, seeds 0 to 29, as in the main text).}}{%
\small\setlength{\tabcolsep}{8pt}
\begin{tabular}{@{}lrrrrr@{}}
\toprule
Sample & $n$ & Masked & Unmasked & Blind & Logged \\
\midrule
All decisions (main text) & 19,827 & 1.4 & 1.5 & 2.3 & 49.7 \\
Without closed other content & 19,218 & 1.0 & 1.0 & 1.8 & 51.4 \\
Earlier message observed & 11,301 & 0.8 & 0.9 & 2.3 & 35.3 \\
\bottomrule
\end{tabular}}
\end{table*}

\begin{figure*}[htbp]
\floatconts{fig:km}{\caption{Kaplan-Meier estimates of the proportion of conversations not yet followed by another, by conversation number, main export (observed), with pointwise 95\% bands; censored at the end of the export. $n$ counts conversations, which are runs of the mother's own messages split at gaps over 12 hours.}}{\includegraphics[width=0.62\textwidth]{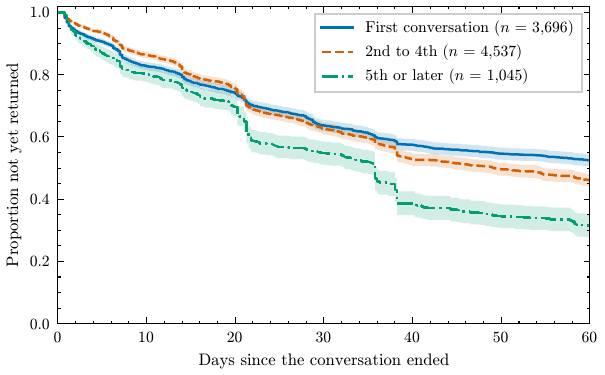}}
\end{figure*}

\begin{figure*}[htbp]
\floatconts{fig:map}{\caption{When waiting beats a template (modelled, main export). (a) Value of waiting up to 7 days minus the value of a template, $V_D - p_T$ ($p_T = 0.81$, $c = 0$), from \propositionref{prop:wait}; diverging scale centred at zero with symmetric limits; black curve: break-even line; dashed lines: urgent (6 h) and routine (48 h) half-lives. (b) Estimated return rates of the closed-window decisions. (c) Share of closed-window decisions for which the threshold rule waits and (d) its gain over the template rule, by template cost $c$ and half-life; waiting becomes common only once $c$ reaches 0.75 to 0.85.}}{\includegraphics[width=\textwidth]{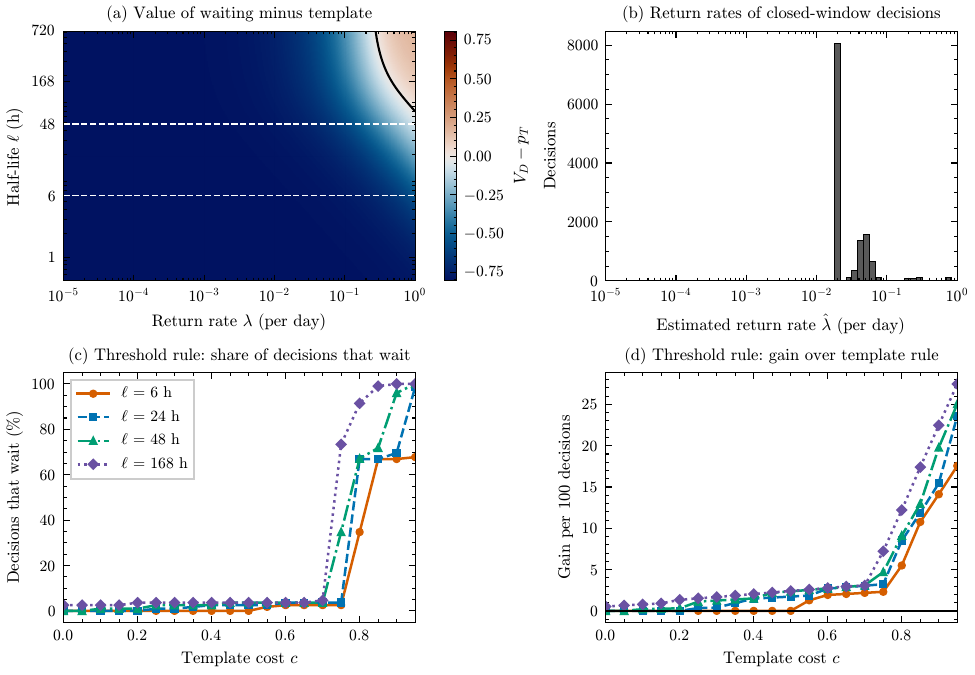}}
\end{figure*}

\begin{figure*}[htbp]
\floatconts{fig:bandit}{\caption{Linear Thompson-sampling learners replayed on the 19,827 logged decisions in time order, main export (modelled rewards). (a) Cumulative regret against the window-aware oracle, mean and 95\% band over 30 runs (seeds 0 to 29). (b) Regret per 100 decisions, log scale, with the logged behaviour.}}{\includegraphics[width=\textwidth]{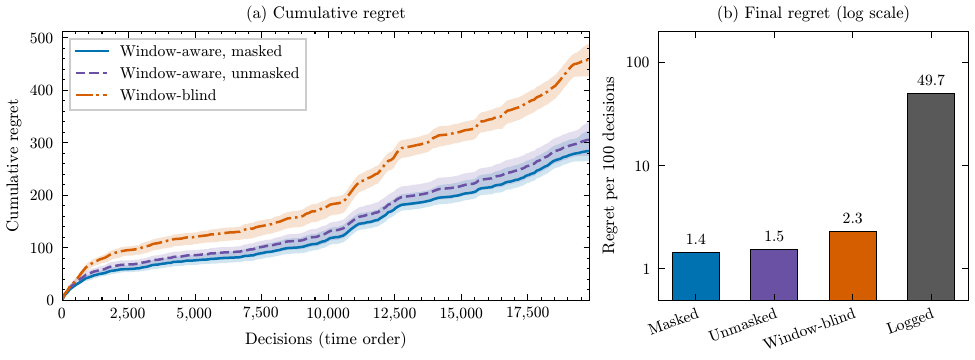}}
\end{figure*}

\section{Simulator}
\label{app:sim}

\paragraph{Inputs.} Gap mixture: quick component share 0.46, means 24.9 and 340 days. Template delivery for the guide: 90.5\% (earlier message observed) and 86.9\% (not); confirmed-only 87.6\% and 78.0\%. Genuine 24-hour reply rate to delivered button templates 17.9\% (11.3\% of raw replies were opt-out taps) and 3.4\% to delivered plain guides; the reply lag is log-normal with log-hours mean 0.59 and standard deviation 1.99, truncated at 24 hours. Opt-out rate 785 strict opt-outs per 46,394 delivered bot-initiated messages. Evaluation: 8 seeds of 3,000 mothers (seeds 1000 to 1007); stress tests 4 seeds of 2,000, with Q-learning retrained on 100,000 mothers.

\paragraph{Dynamics.} A reply is a conversation: it opens the window for 24 hours and restarts the mother's renewal process, so her next spontaneous message follows a fresh gap. With weekly decisions, a reply in one week rarely leaves the window open at the next. In the persistence stress test, each mother draws her own probability of a quick gap from a Beta distribution with mean 0.46 and concentration $\kappa = 1$, instead of all mothers sharing 0.46.

\paragraph{Q-learning.} The state is the band of hours since her last message (none, 0 to 24, 24 to 72, 72 to 168, over 168) and the number of conversations in the last 28 days (0, 1, 2 or more), 15 states in all. The actions are $F$, $T$, wait, $T^+$ and hybrid, with $F$ allowed only when the window is open; there is no option to withhold the offer in the main runs. The Q-table starts at zero. The weekly reward is the guide delivered plus the offer's discounted value, minus 0.05 for every paid message, template or service; the stress tests labelled ``cost per paid message'' change this 0.05. The main run trains on 150,000 simulated mothers and each stress test on 100,000, enough because the greedy policy stops changing after about 10,000 (\figureref{fig:stress}b). The reply-button strategy is left out of the stress tests to save computation; in the main setting it delivers fewer offers and less reward than the hybrid, and \tableref{tab:suppress} includes it. \emph{Value} in \tableref{tab:strategies} is discounted offers per 100 due; \emph{reward} is the undiscounted sum over 26 weeks of the weekly reward, so the template rule's 44.6 is about 23.5 guides plus 23.5 offers less the message costs.

\paragraph{Suppressing the offer.} \tableref{tab:suppress} adds a ``none'' action and raises the per-message exit probability of the global model in steps. Q-learning (60,000 training mothers; 4 seeds of 2,000) matches the template rule within simulation noise up to four times the observed bound; at eight and sixteen times it suppresses more offers, delivering 71.9 and 46.5 per 100, and gains at most 0.2 reward units.

\paragraph{Results.} \figureref{fig:val} compares the simulator with the logs, \tableref{tab:strategies} and \figureref{fig:cost} give each strategy's value and cost, and \figureref{fig:stress} gives the stress tests and the Q-learning training curve.

\begin{table}[h]
\floatconts{tab:strategies}{\caption{Simulated strategies over 26 weeks, main setting (stream-specific opt-out), means over 8 seeds of 3,000 mothers. Offers delivered and discounted value per 100 offers due; reward and cost per mother. Cost counts templates and in-window service messages at US\$0.0101 each, as under pricing from October 2026.}}{%
\small
\begin{tabular}{@{}lrrrr@{}}
\toprule
Strategy & Offers & Value & Reward & Cost \\
\midrule
Logged & 1.3 & 1.3 & 22.5 & 0.24 \\
Wait rule & 10.2 & 3.8 & 23.1 & 0.26 \\
Reply button ($T^+$) & 18.5 & 13.8 & 25.6 & 0.29 \\
Hybrid & 91.7 & 51.2 & 34.4 & 0.48 \\
Template rule & 90.3 & 90.3 & 44.6 & 0.47 \\
Threshold & 90.3 & 90.3 & 44.6 & 0.47 \\
Q-learning & 90.3 & 90.3 & 44.6 & 0.47 \\
\bottomrule
\end{tabular}}
\end{table}

\begin{table}[h]
\floatconts{tab:suppress}{\caption{Simulated reward per mother over 26 weeks when every delivered message ends contact with probability $\rho$ (global opt-out model) and Q-learning may suppress the offer. $\rho = 1.7\%$ is the observed upper bound. Last column: offers Q-learning delivers per 100 due. Means over 4 seeds of 2,000 mothers.}}{%
\small\setlength{\tabcolsep}{3pt}
\begin{tabular}{@{}rrrrrr@{}}
\toprule
$\rho$ & Template & Suppress & $T^+$ & Q-learning & Offers \\
\midrule
1.7\% & 30.8 & 18.7 & 20.6 & 30.8 & 89.0 \\
3.4\% & 22.4 & 15.7 & 16.7 & 22.3 & 87.2 \\
6.8\% & 13.4 & 11.3 & 11.5 & 13.5 & 83.6 \\
13.5\% & 7.0 & 6.8 & 6.6 & 7.1 & 71.9 \\
27.1\% & 3.5 & 3.5 & 3.4 & 3.5 & 46.5 \\
\bottomrule
\end{tabular}}
\end{table}

\begin{figure*}[htbp]
\floatconts{fig:val}{\caption{Simulator validation under the logged behaviour. (a) Proportion of mothers who have written again, by days since a conversation: simulated, the fitted two-component mixture, and Kaplan-Meier estimates from the logs (the points the mixture was fitted to). (b) Statistics in the logs and in the simulation, grouped by role: a calibrated input, targets of the gap-mixture fit, and checks not used in fitting. The simulation overstates returns at 30 and 90 days (43.2\% against 37.8\%, 69.3\% against 58.5\%). Agreement is conditional on the calibrated model and does not validate its behavioural assumptions.}}{\includegraphics[width=\textwidth]{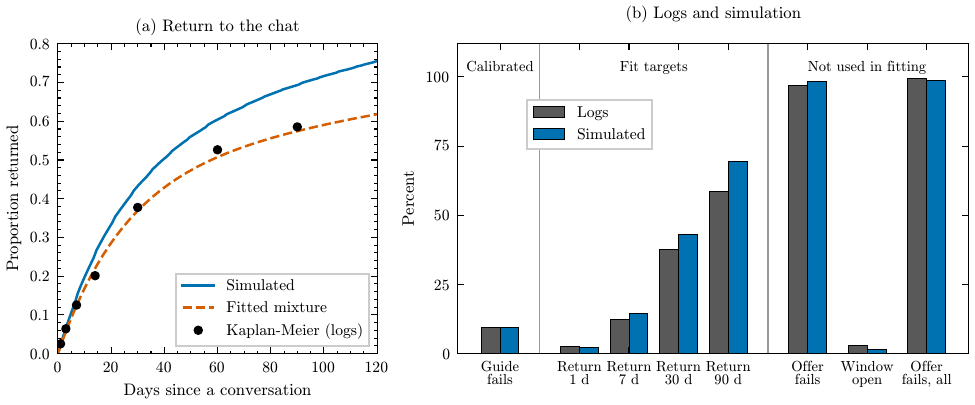}}
\end{figure*}

\begin{figure*}[htbp]
\floatconts{fig:cost}{\caption{Simulated value against cost per mother over 26 weeks, main setting. Value is discounted offers delivered per 100 due; cost counts templates and in-window service messages at US\$0.0101 each, as under pricing from October 2026. Means with 95\% intervals over 8 seeds, narrower than the markers. The template rule, the threshold rule and Q-learning fall on the same point.}}{\includegraphics[width=0.6\textwidth]{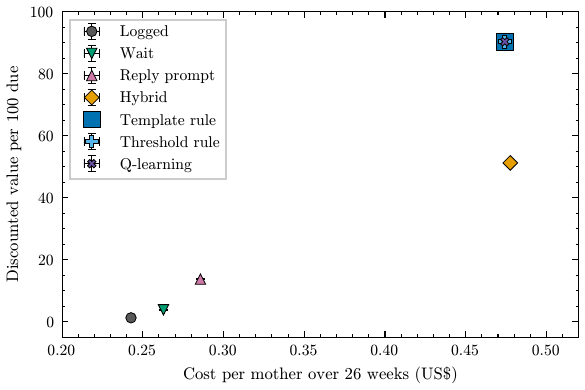}}
\end{figure*}

\begin{figure*}[htbp]
\floatconts{fig:stress}{\caption{Stress tests (simulated). (a) Reward per mother of each strategy minus the template rule, in the main setting and 13 single-input stress tests; 4 seeds of 2,000 mothers, Q-learning retrained for each test. One-sided colour scale from zero (white) to the largest loss; values rounded to 0.1, so gaps below 0.05 show as 0.0 (the largest standard error of a reward is 0.19). (b) Reward of Q-learning's greedy policy by number of training mothers, with the template rule (line and 95\% band) and the logged behaviour.}}{\includegraphics[width=\textwidth]{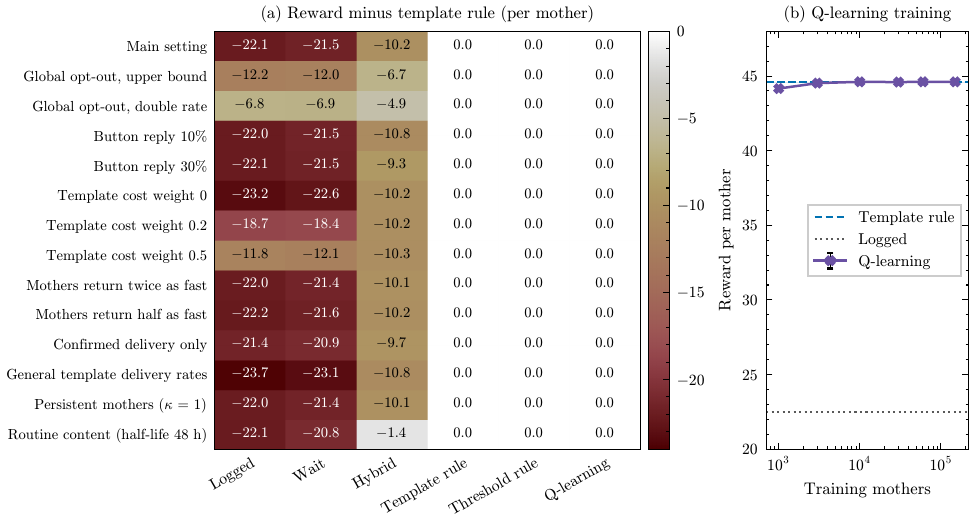}}
\end{figure*}

\section{Burden Models}
\label{app:burden}

\paragraph{Definitions.} The unit is a delivered (not failed) bot-initiated message: 49,256 in the main export, of which 46,394 were confirmed delivered and 2,862 had status ``sent'' or ``unknown''; the simulator's opt-out rate uses the confirmed count from the raw export as its denominator. Exposure is $\log(1 + m)$, where $m$ is the number of such messages, or of templates, in the previous 7 days (24 hours in one specification), excluding the same chain (10 minutes). An increase of $\ln 2 = 0.69$ log units doubles $1 + m$, so ``per doubling'' means doubling one plus the count. Outcomes are an opt-out within 24 hours (a message starting ``opt out'' or exactly ``stop'' or ``unsubscribe'') and reading within 24 hours. Opt-out models use only flows in which an opt-out occurred. ``Recency'' is a set of indicators for hours since the mother's last message (24 or fewer, 1 to 7 days, 7 to 30 days, over 30 days or none observed).

\paragraph{Models.} Within-mother models subtract each mother's mean (mothers with at least 3 messages for opting out, 5 for reading) and add flow and month effects. Because opting out ends some streams, a mother who opts out early has lower later volume, which biases within-mother coefficients downward. The pooled discrete-time model avoids this: it keeps each mother's messages until her first opt-out, models whether that opt-out follows within 24 hours by logistic regression on template volume, recency, $\log(1 + \text{days since her first message})$, flow and month, and does not demean. Errors are clustered by mother throughout. \tableref{tab:burden} gives every specification.

\begin{table}[h]
\floatconts{tab:burden}{\caption{Burden models, main export (observed associations), with 95\% intervals clustered by mother. Top: within-mother linear models, percentage points per log unit of exposure. Bottom: pooled logistic models, odds ratio per log unit. $^a$Template volume, recency and tenure.}}{%
\small\setlength{\tabcolsep}{3pt}
\begin{tabular}{@{}lr@{}}
\toprule
Specification & Estimate (95\% CI) \\
\midrule
\multicolumn{2}{@{}l}{\emph{Opt-out within 24 hours, within mother}} \\
All messages, 7 days & $-2.6$ ($-3.4$ to $-1.9$) \\
All messages, 24 hours & $-2.8$ ($-3.7$ to $-1.9$) \\
Before first opt-out only & $-1.4$ ($-2.1$ to $-0.7$) \\
Templates, with recency & $-2.3$ ($-3.5$ to $-1.2$) \\
\multicolumn{2}{@{}l}{\emph{Read within 24 hours, within mother}} \\
All messages, 7 days & $-2.5$ ($-3.4$ to $-1.5$) \\
Templates, with recency & $-3.2$ ($-4.4$ to $-2.0$) \\
\multicolumn{2}{@{}l}{\emph{Opt-out within 24 hours, pooled (odds ratio)}} \\
All messages, 7 days & 0.69 (0.61 to 0.78) \\
First opt-out, with controls$^a$ & 0.94 (0.75 to 1.17) \\
\bottomrule
\end{tabular}}
\end{table}

\section{Send Time}
\label{app:sendtime}

The sample is 17 batched broadcasts (7,794 delivered messages, 2,133 mothers). Windows are morning, afternoon, evening and night (22:00 to 23:59), compared with campaign fixed effects. The night group is small: 266 messages, 193 (73\%) from one campaign, so 73 remain when that campaign is dropped. Intervals are clustered by mother; clustering by the 17 campaigns instead gives a narrower interval for the night difference ($-26.2$ to $-19.8$), which we do not rely on with so few clusters. Read within 6 hours, night minus morning: $-23.0$ points (95\% CI $-30.9$ to $-15.2$); the intervals for reading within 1 hour and reading eventually include zero. Afternoon and evening do not differ from morning on any outcome. Dropping each campaign in turn gives $-19.5$ to $-23.8$ points. Prior reading and recent contact are balanced across send windows (joint $p = 0.86$ and $0.36$). \figureref{fig:time} shows all estimates.

\begin{figure*}[htbp]
\floatconts{fig:time}{\caption{Reading by send time within 17 batched broadcasts (7,794 delivered messages), main export. Difference from morning sends, in percentage points, in the share read eventually, within 6 hours and within 1 hour, with campaign fixed effects and 95\% intervals clustered by mother. An observed association, not a randomised effect: the order of sending within a batch is undocumented.}}{\includegraphics[width=0.6\textwidth]{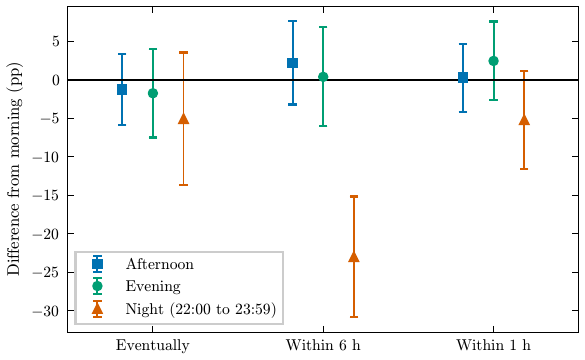}}
\end{figure*}

\section{Replication}
\label{app:replication}

The replication export was analysed with unchanged code. \tableref{tab:replication} sets its results beside the main export. The jump at 24 hours, the share of failures from the same two flows and the ranking of the options for the medical offer all replicate. Template values and waiting are lower in the replication export because more of its closed-window decisions have no earlier message observed (2,072 of 2,439, 85\%, against 8,526 of 12,751, 67\%), and that group includes numbers no longer on WhatsApp. With an earlier message observed, the template reaches 95.9 to 98.5 per 100 offers in the replication export (123 decisions) and 97.3 to 99.0 in the main export. The threshold rule waits on 0.5\% of decisions instead of none.

\begin{table*}[h]
\floatconts{tab:replication}{\caption{Main and replication exports side by side. Observed results except the template values (modelled; lower bound confirmed delivery, upper bound not failed) and the simulation gap (simulated). Medical-offer values are per 100 closed-window offer decisions ($n$ = 3,933 and 752) with 95\% intervals from 500 mother-cluster bootstrap replicates that do not refit the models. The upper limit of the replication jump exceeds 100 because the linear probability model's intervals are not bounded.}}{%
\small\setlength{\tabcolsep}{6pt}
\begin{tabular}{@{}lrr@{}}
\toprule
 & Main & Replication \\
\midrule
Free-form failed, open & 0 of 35,500 & 0 of 9,546 \\
Free-form failed, closed & 98.6\% & 97.6\% \\
\quad closed messages & 6,267 & 544 \\
18 to 30 h, before 24 h & 0 of 114 & 0 of 19 \\
18 to 30 h, after 24 h & 72 of 79 & 18 of 18 \\
Jump at 24 h (points) & 93.0 & 100.0 \\
\quad 95\% CI & 87.0 to 99.1 & 94.5 to 105.6 \\
Two flows, share of failures & 66.1\% & 52.0\% \\
Offer, logged & 0.3 (0.1 to 0.5) & 0.4 (0.0 to 0.9) \\
Offer, template, lower & 81.5 (80.8 to 82.2) & 69.1 (67.5 to 70.9) \\
Offer, template, upper & 86.8 (86.3 to 87.4) & 77.6 (76.4 to 79.0) \\
Offer, wait 7 days & 16.6 (15.2 to 18.1) & 7.0 (4.9 to 9.7) \\
Offer, reply in 24 h & 11.6 (10.5 to 13.0) & 5.3 (3.3 to 7.7) \\
Threshold rule waits & 0.0\% & 0.5\% \\
Largest Q-learning gap & 0.05 & 0.16 \\
\bottomrule
\end{tabular}}
\end{table*}

\section{Software, Seeds and Tests}
\label{app:repro}

Software: Python 3.13, pandas 3.0, NumPy 2.5, SciPy 1.18, statsmodels 0.15, scikit-learn 1.9, matplotlib 3.11. Seeds: 42 (cross-fitting folds, policy bootstrap, Cox bootstrap), 7 (medical-offer bootstrap), 11 (content-type refit bootstrap and Q-learning training), 0 to 29 (Thompson-sampling runs), 1000 to 1007 (simulation evaluation); for the review checks, 5 (restricted-sample bootstrap), 0 to 9 (restricted replay) and 1000 to 1003 (suppress experiment). One command runs each pipeline end to end; the review checks in Appendices B to G are separate scripts that read its outputs. Unit tests cover the window, delivery and decision functions, the simulator and the result tables, and an automated check compares every number quoted in this paper with the result tables.

\end{document}